%% file: template.tex
\documentclass[14pt]{article}

\usepackage{arxiv}
\usepackage[utf8]{inputenc} 
\usepackage[T1]{fontenc}    

\usepackage{url}            
\usepackage{booktabs}       
\usepackage{amsfonts}       
\usepackage{nicefrac}       
\usepackage{microtype}      
\usepackage{lipsum}
\usepackage{graphicx}
\graphicspath{ {./images/} }
\input{math_commands.tex}

\usepackage{pgfplots}
\usepackage{pgfplotstable}
\pgfplotsset{compat=1.18}
\usepackage{hyperref}
\usepackage{url}
\usepackage{graphicx}
\usepackage{colortbl}
\usepackage{xcolor}
\usepackage[table]{xcolor}
\usepackage{xfp}
\usepackage{enumitem}
\usepackage{siunitx}
\usepackage[numbers,sort&compress]{natbib}
\usepackage{amsmath,amssymb,amsthm}
\usepackage[ruled,vlined]{algorithm2e}
\usepackage{amsmath,amssymb}
\usepackage{longtable}

\newtheorem{theorem}{Theorem}  
\newtheorem{lemma}{Lemma}

\makeatletter
\newlength{\authorcolsep}
\newlength{\authorrowsep}

\newcommand{\Author}[3]{%
  \begin{tabular}[t]{@{}c@{}}
    {\large\bfseries #1}\\[0.2em]
    {\small #2}\\[-0.15em]
    {\small\ttfamily #3} 
  \end{tabular}%
}
\newcommand{\eq}{\textsuperscript{\dag}}  
\newcommand{\corr}{\textsuperscript{*}}   
\makeatother
\title{ResT: Reshaping Token-Level Policy Gradients for Tool-Use Large Language Models}
\author{%
\begin{tabular}[t]{@{}c@{\hspace{\authorcolsep}}c@{\hspace{\authorcolsep}}c@{}}
\Author{Zihan Lin\eq}{CRIPAC, Institute of Automation}{zihan.lin@nlpr.ia.ac.cn} &
\Author{Xiaohan Wang\eq}{Meituan}{wangxiaohan17@meituan.com} &
\Author{Jie Cao}{CRIPAC, Institute of Automation}{jie.cao@cripac.ia.ac.cn} \\[\authorrowsep]
\Author{Jiajun Chai}{Meituan}{chaijiajun@meituan.com} &
\Author{Guojun Yin\corr}{Meituan}{yinguojun02@meituan.com} &
\Author{Wei Lin}{Meituan}{linwei31@meituan.com} \\[\authorrowsep]
\multicolumn{3}{c}{\Author{Ran He}{CRIPAC, Institute of Automation}{rhe@nlpr.ia.ac.cn}} \\[0.6em]
\multicolumn{3}{c}{\small \eq~Equal contribution \quad \corr~Corresponding author}
\end{tabular}
}

\begin{document}
\maketitle
\begin{abstract}
Large language models (LLMs) transcend passive generation and act as goal-directed agents by invoking external tools. Reinforcement learning (RL) offers a principled framework for optimizing these emergent tool-use policies, yet the prevailing paradigm relies exclusively on sparse outcome rewards and lacks consideration of the particularity of tool-use tasks, inflating policy-gradient variance and resulting in inefficient training.
To better understand and address these challenges, we first establish a theoretical link between policy entropy and training stability of tool-use tasks, which reveals that structured, low-entropy tokens are primary determinants of rewards. Motivated by this insight, we propose \textbf{Res}haped \textbf{T}oken-level policy gradients (\textbf{ResT}) for tool-use tasks. ResT reshapes the policy gradient through entropy-informed token reweighting, progressively upweighting reasoning tokens as training proceeds. This entropy-aware scheme enables a smooth shift from structural correctness to semantic reasoning and stabilizes convergence in multi-turn tool-use tasks. Evaluation on BFCL and API-Bank shows that ResT achieves state-of-the-art results, outperforming prior methods by up to $8.76\%$. When fine-tuned on a 4B base LLM, ResT further surpasses GPT-4o by $4.11\%$ on single-turn tasks and $1.50\%$ on multi-turn base tasks.
\end{abstract}


\section{Introduction}

The recent emergence of large language models (LLMs) has precipitated a paradigm shift from passive, data-driven models to proactive, goal-oriented autonomous agents with complex tool-use abilities \citep{cheng2024exploring, xi2025survey}. Consequently, considerable research has focused on post-training LLMs to enhance their tool-use capabilities for domain-specific applications such as deep search and travel assistance. \citep{wang2025toward, feng2025retool}. Post-training for agentic LLMs centers on supervised fine-tuning (SFT) and Reinforcement learning (RL) with distinct trade-offs \citep{chu2025sft, wang2025parameter}. In tool-use tasks, RL demonstrates strong generalization and robust performance \citep{song2025r1, qian2025toolrl, feng2025group}. Most approaches, however, assign a single outcome reward after multi-turn tool calls \citep{jin2025search}; despite gains in multi-hop search \citep{liu2025omne}, this paradigm introduces two challenges for general tool use: First, reward design is intrinsically noisy. Unlike multi-hop search with near-unique references, many practical tool-use tasks admit multiple valid outputs (e.g., recommendations). As a result, outcome-only rewards induce high-variance gradients and provide weak incentives for reasoning, even when augmented with LLM-as-a-judge or learned reward models \citep{wang2025ragen, gu2025surveyllmasajudge}.
Second, multi-turn tool-use tasks are systems-inefficient: per-sample interactions, concurrency control, and variable horizons depress throughput and inflate infrastructure cost, making tool-use RL substantially heavier than SFT and single-turn RL \citep{team2025kimi}.

The challenges motivate a central question: Could a tool-use agent be trained by rewarding each turn with specific and low-variance rewards? Inspired by Deepseek-R1’s use of GRPO with rule-based rewards, we compute rule-based matching scores between generated responses and ground truth. Recent work, such as ToolRL \citep{qian2025toolrl}, has validated this approach, achieving SOTA performance. Specifically, these studies decompose multi-turn tool calls into a series of single-turn tasks and establish rule-based matching rewards for efficient RL training. However, prior work often overlooks a crucial aspect: in tool-call settings, rule-based rewards are inherently concentrated on format tags, tool names, and key parameters, while reasoning and general-response tokens contribute little during the early stages of training. Uniformly treating all tokens dilutes the RL signal and is particularly detrimental for approaches lacking token-level critics (e.g., GRPO). 
\begin{figure}[h]
    \centering
    \includegraphics[width=0.99\textwidth]{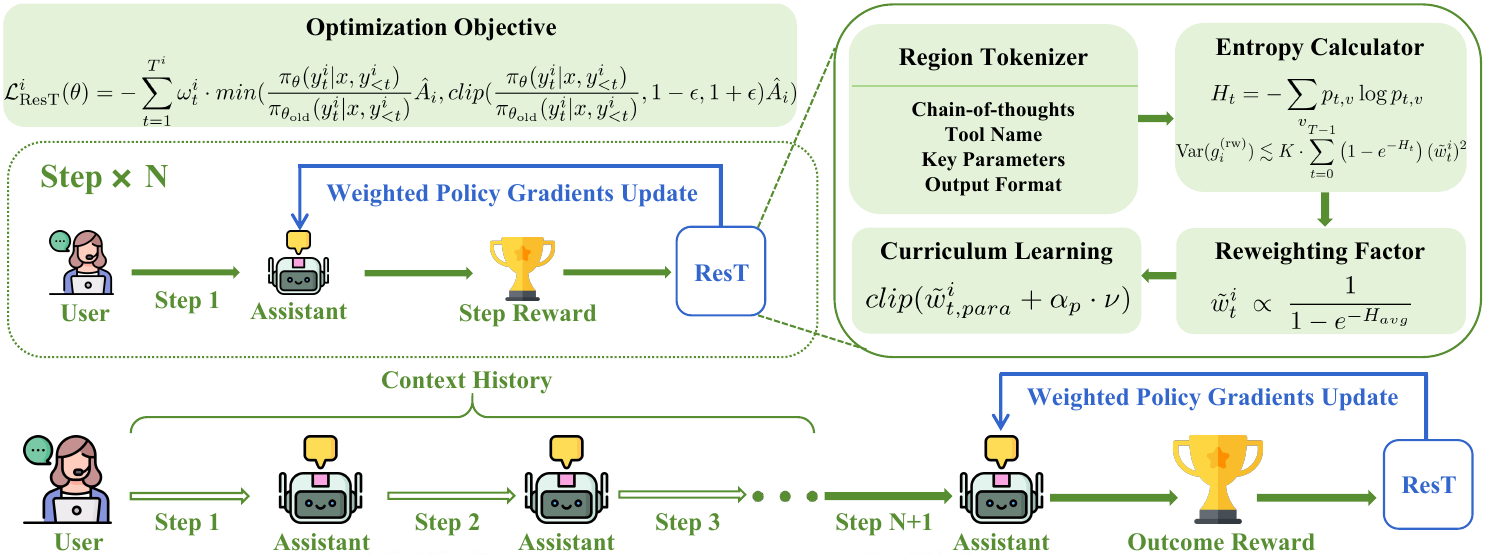}
\caption{ ResT decomposes multi-turn tool-use tasks into single-turn tasks and further reshapes the policy gradient according to the average entropy in different regions, enabling dense and effective reward signals.}
    \label{fig:overview}
    \vspace{-1.0em}
\end{figure}

To effectively and efficiently train tool-use agents, this paper proposes to \textbf{Res}hape \textbf{T}oken-level policy gradients based on single-turn RL training (\textbf{ResT}). As introduced in Figure \ref{fig:overview}, ResT factorizes multi-turn tool use into single-turn steps and reshapes policy gradients with region-level average entropy. This design emphasizes structured, low-entropy tokens such as tool names and parameters, yielding dense and effective reward signals. While RL for tool-use agents should ideally progress from format compliance to parameter accuracy and ultimately complex reasoning, such curricula are seldom adopted. In ResT, this progression is guided by adaptive weighting across token regions, providing a lightweight curriculum that further stabilizes training and supports state-of-the-art results on tool-use benchmarks. The contributions are summarized as follows:

1) We establish a theoretical and empirical link between policy entropy and training stability. Specifically, we demonstrate that a lower average entropy correlates with reduced variance in policy-gradient updates. This insight reveals that structured, low-entropy tokens—such as tool names and parameters—are primary determinants of the resulting rewards. By uncovering this connection, we provide a principled explanation for why emphasizing these tokens during training leads to more stable optimization and improved performance on tool-use tasks.

2) We introduce an entropy-aware, token-level reshaping mechanism that incorporates curriculum learning. Within this framework, tool-use and reasoning tokens are weighted by their region-level average entropy, which directly influences training stability and regulates how different token categories contribute to reward signals. As training progresses and entropy decreases, the weights for reasoning tokens are gradually increased. This curriculum-based approach stabilizes convergence and systematically enhances the model's reasoning capabilities in complex, multi-turn tool-calling scenarios.

3) Experimental results across multiple benchmark datasets show that our proposed method improves upon existing approaches by up to $8.76\%$, establishing a new state-of-the-art. Notably, when fine-tuned on the Qwen3-4B-2507 base model, our method outperforms GPT-4o \citep{hurst2024gpt} by $1.50\%$ on multi-turn tool-use base tasks and $4.11\%$ on single-turn tool-use tasks. Furthermore, comprehensive ablation studies reveal that our curriculum-based reshaping outperforms static reward weighting by up to $4.86\%$. 


\section{Related Work}
\subsection{Reinforcement Learning for LLM Reasoning}
Research on reinforcement learning (RL) for large language models (LLMs) progressed from early paradigms such as reinforcement learning from human feedback (RLHF) \citep{kaufmann2024survey} and proximal policy optimization (PPO) \citep{schulman2017proximal} to a class of methods based on preferences and policies, including direct preference optimization (DPO) \citep{rafailov2023direct}, SimPO \citep{meng2024simpo}, and group relative policy optimization (GRPO) \citep{shao2024deepseekmathpushinglimitsmathematical}. Subsequent variants, such as dynamic sampling policy optimization (DAPO) \citep{yu2025dapoopensourcellmreinforcement} and value-augmented PPO (VAPO) \citep{yue2025vapo}, emphasize training stability and sample efficiency. Taken together, these advances indicate that RL can substantively enhance LLM reasoning, moving beyond surface-level pattern matching toward competencies such as logical inference and strategic planning \citep{deepseekai2025deepseekr1incentivizingreasoningcapability,kimiteam2025kimik15scalingreinforcement}.

\subsection{Tool-Use Agents}
Tool use extends LLMs from passive text generators to interactive systems capable of invoking external tools and APIs \citep{zhang2025policy,hao2025funreasonenhancinglargelanguage}. Existing approaches span prompt engineering and supervised fine-tuning (SFT) \citep{chen2023sft} on curated corpora \citep{qin-etal-2025-meta,hao2025funreasonenhancinglargelanguage,zhang2024xlamfamilylargeaction}, alongside specialized architectures and retrieval-augmented pipelines that select tools conditioned on the query \citep{nguyen2024sfrragcontextuallyfaithfulllms}. Recent agentic frameworks further support multi-step planning and autonomous tool orchestration \citep{team2025kimi}. However, most methods remain predominantly supervised, limiting exploration of diverse reasoning strategies and generalization to novel tool-use scenarios \citep{liu2025toolace}.

To mitigate these limitations, RL formulations have been explored for tool-using agents. ToolRL \citep{qian2025toolrl} systematically compares SFT with GRPO \citep{shao2024deepseekmathpushinglimitsmathematical}; GIGPO \citep{feng2025group} adapts GRPO to multi-step tool use via multi-trajectory sampling and a two-level reward scheme; PVPO \citep{feng2025pvpo} further reduces rollout dependence and alleviates local optima through a reference–anchor mechanism and data pre-sampling; meanwhile, Agent Lightning \citep{luo2025agentlightningtrainai} decouples training from execution and employs hierarchical RL to convert arbitrary agent trajectories into transition-level learning signals. Despite these advances, practical RL methods still largely optimize at the sequence level, where coarse-grained rewards induce high variance in policy gradients, constraining reward granularity and impeding efficient credit assignment and optimization.

\begin{figure}[h]  
    \centering
    \includegraphics[width=0.99\textwidth]{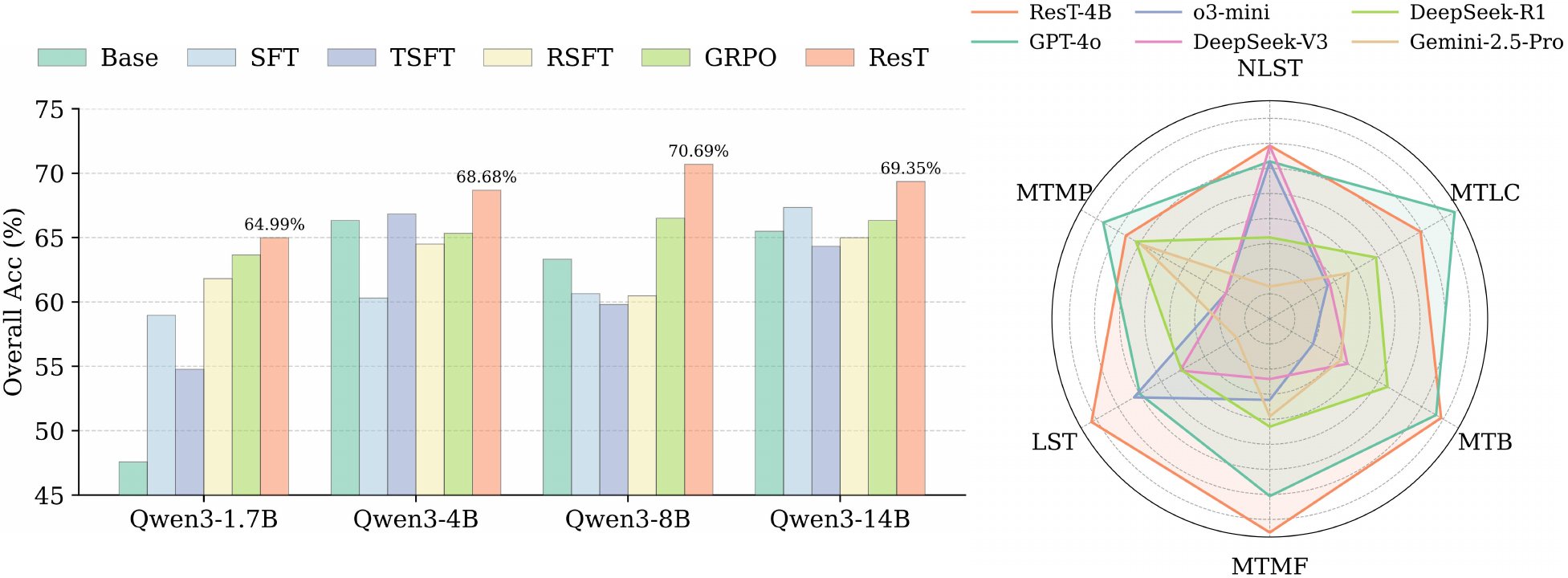}
\caption{\textbf{Left:} Overall accuracy on the API-Bank test set. \textbf{Right:} Overall accuracy on the BFCL test set. Axes are: NLST: None-Live Single Turn, MTLC: Multi-Turn Long Context, MTB: Multi-Turn Base, MTMF: Multi-Turn w/ Missing Functions, LST: Live Single Turn, MTMP: Multi-Turn w/ Missing Parameters. }
    \label{fig:result}
    \vspace{-1.0em}
\end{figure}
\section{Method}
\label{method}
In this section, we introduce two key innovations. First, we prove that lower average token entropy reduces policy-gradient variance and show that structured low-entropy tokens dominate the reward signal. Second, we propose token-level gradient reshaping, along with curriculum learning, to amplify updates for reward-critical tokens (e.g., tool names, arguments, output format), thereby mitigating variance and improving performance on complex tool-use tasks.

\subsection{Theoretical Motivation}
\label{theory}
Successful tool-use hinges on generating a few critical tokens for names, parameters, and output formats. Conventional uniform rewards obscure these sparse signals, leading to high-variance gradients and unstable convergence. To address this, we introduce a theoretical framework for reducing policy gradient variance via entropy-aware token reweighting. Our analysis decomposes the gradient variance, links it to token-level entropy, and yields an optimal reweighting scheme that markedly improves training efficiency for high-precision tasks.

\begin{lemma}[Policy Gradient Variance Decomposition] 
\label{lem:var-decomp} 
Let $J(\theta)=\E_{\tau\sim\pi_\theta}[R(\tau)]$ denote the expected return. For a single trajectory $\tau_i=(y_{i,0:T-1})$, define the trajectory-level gradient 
\begin{equation} 
 g_i \;=\;\Biggl(\sum_{t=0}^{T-1}\nabla_\theta \log \pi_\theta(y_{i,t}\mid x,y_{i,<t})\Biggr)\,\hat A_i,  
\end{equation} 
where $\hat A_i$ is advantage function. Given $G$ i.i.d.\ trajectories $\{\tau_i\}_{i=1}^G$ from $\pi_\theta$, the mini-batch estimator is $\widehat{\nabla J}=(1/G)\sum_{i=1}^G g_i$. Then for each coordinate $k$ the variance satisfies
\begin{equation} 
\label{eq:var-mean} 
\operatorname{Var}(\widehat{\nabla J}^{(k)})=\tfrac{1}{G}\operatorname{Var}(g_i^{(k)}), 
\end{equation} 
If $\hat A_i=(r_i-\mu)/\sigma$ with $\mu=\E[r]$ and $\sigma^2=\Var(r)$, then $\E[g_i]=\nabla_\theta J(\theta)$ and $\widehat{\nabla J}$ is unbiased. Proof is deferred to Appendix~\ref{appendix:proof_lemma_var_decomp}. \end{lemma}

Lemma~\ref{lem:var-decomp} establishes the variance scaling law under mini-batch sampling. Building on this decomposition, we next characterize the second-order moment of $g_i$ and connect it to token-level entropy.

\begin{lemma}[Second-Order Moment and Entropy Connection]
\label{prop:moment-entropy}
For the single-trajectory estimator $g_i$, the variance satisfies
\begin{equation}
\label{eq:prop-moment}
\Var(g_i)\;\le\;\E\!\left[\hat A_i^{\,2}\,
\Biggl\lVert \sum_{t=0}^{T-1} J_t^\top s_t \Biggr\rVert^2\right],
\end{equation}
where $s_t=e_{y_t}-p_t$ is the logit-space score, $p_t=\softmax(z_t)$ is the token distribution with $z_t$ being the logits, and $J_t=\partial z_t/\partial\theta$ is the Jacobian.  
Moreover, the per-step score satisfies
\begin{equation}
\label{eq:prop-entropy}
\E[\|s_t\|^2]=1-\sum_v p_{t,v}^2 \;\le\; 1-e^{-H_t},
\end{equation}
where $H_t=-\sum_v p_{t,v}\log p_{t,v}$ is the Shannon entropy, with $v$ indexing over the token vocabulary. Proof is deferred to Appendix~\ref{appendix:proof_prop_moment_entropy}.
\end{lemma}

Lemma~\ref{prop:moment-entropy} shows that variance fundamentally depends on both squared advantage and token entropy. This result provides a natural handle for reweighting gradient contributions across timesteps. For a sequence of length $T$, we denote the reweighted policy gradient estimator as:
\begin{equation}
\label{eq:estimator}
g_i^{(\mathrm{rw})}
:= \Biggl(\sum_{t=0}^{T-1} \tilde{w}_t \nabla_\theta\log \pi_\theta(y_{i,t}\mid x,y_{i,<t})\Biggr)\hat A_i,
\qquad \text{subject to} \quad \sum_{t=0}^{T-1} \tilde{w}_t=T,
\end{equation}
\begin{theorem}[Variance Upper Bound for Entropy-Aware Reweighting]
\label{thm:main}
The variance of the reweighted estimator $g_i^{(\mathrm{rw})}$ is upper-bounded using Lemma \ref{prop:moment-entropy}:
\begin{equation}
\label{eq:thm-upper}
\Var\!\big(g_i^{(\mathrm{rw})}\big)
\;\le\; \E[\hat A_i^{\,2}] \sum_{t=0}^{T-1}\beta_t \,(\tilde{w}_t)^2,
\end{equation}
where $\beta_t=\E[\|J_t\|_F^2(1-e^{-H_t})]$ quantifies the variance contribution of step $t$ by combining the Jacobian norm and the token entropy.
\end{theorem}

Building on this upper bound, we next seek the optimal reweighting scheme that minimizes the variance. This leads to a closed-form characterization of the weights.

\begin{theorem}[Optimal Entropy-Aware Reweighting]
\label{thm:opt-weights}
Minimizing the variance bound in Theorem \ref{thm:main} with respect to the weights $\tilde{w}_t$ yields a closed-form solution: (Proof is deferred to Appendix~\ref{appendix:proof_main}.)
\begin{equation}
\label{eq:optimal-w}
\tilde{w}_t^{\star}=\frac{T}{\sum_{u=0}^{T-1}\beta_u^{-1}}\cdot \frac{1}{\beta_t},
\end{equation}
which effectively down-weights sequence positions with larger intrinsic variance contributions. Substituting the optimal weights $\tilde{w}_t^\star$ into Theorem \ref{thm:main} gives the minimized variance bound:
\begin{equation}
\label{eq:thm-min}
\Var\!\big(g_i^{(\mathrm{rw})}\big)
\;\le\; \E[\hat A_i^{\,2}] \cdot \frac{T^2}{\sum_{t=0}^{T-1}\beta_t^{-1}}.
\end{equation}
\end{theorem}

In practice, computing $\beta_t$ exactly may be computationally expensive.  
A practical alternative is to approximate $\beta_t$ using entropy-only statistics, which gives simple weighting rules such as
\begin{equation}
\label{factor}
\tilde{w}_t\;\propto\;\frac{1}{1-e^{-H_{avg}}}
\quad\text{or}\quad
\tilde{w}_t\;\propto\;\frac{1}{H_{avg}}.
\end{equation}
$H_{avg}$ denotes the average entropy computed separately for each region (format tags, tool names, key parameters, and chain-of-thoughts) in tool-use tasks. These approximations preserve the inverse relationship between entropy and variance contribution, and thus serve as efficient surrogates for the optimal weights in Equation~\ref{eq:optimal-w}.

Our findings lead to a principled variance-reduction methodology for policy-gradient methods in tool-use tasks: open-ended tokens (e.g., chain-of-thought), characterized by higher entropy \citep{besta2025reasoninglanguagemodelsblueprint} (see Figure \ref{fig:entropy}), are down-weighted to tighten variance bounds, improve sample efficiency, and preserve estimator unbiasedness.

\subsection{Algorithm Design}

\subsubsection{Reward Score}
Rule-based reward mechanisms have been empirically validated to achieve robust performance across diverse settings. In line with prior work \citep{qian2025toolrl,jin2025search,liao2024mariomathreasoningcode,xie2025logicrlunleashingllmreasoning}, our training objective incorporates components for both output format and execution correctness. Accordingly, we formulate the total reward as a weighted sum of a format matching score and a tool-calling correctness score. Format score of the $i$-th sample is determined by exact-match:
\begin{equation}
\mathcal{S}_{i,\text{format}} = 
\begin{cases} 
1, & \text{if all required fields are complete in order} \\ 
0, & \text{otherwise}
\end{cases}
\end{equation}
The tool-calling correctness score is based on three components: tool name, parameter names, and parameter values. We measure name accuracy using the Jaccard similarity coefficient, defined as the intersection over union of the predicted \(P\) and ground-truth \(G\) sets:
\begin{equation}
    r_{\text{name}} = \frac{|N_G \cap N_P|}{|N_G \cup N_P|} \in [0,1], \quad r_{\text{para}} = \sum_{G_i \in G} \frac{|P_G \cap P_P|}{|P_G \cup P_P|} \in [0, |G|],
\end{equation}
where \(N_G\) and \(N_P\) are the sets of ground-truth and predicted tool names, while \(P_G\) and \(P_P\) the corresponding sets of parameter names for a given tool. Due to the precision requirements of parameter values, we use an exact-match criterion for their evaluation:
\begin{equation}
    r_{\text{value}} = \sum_{G_i \in G} \sum_{v \in \textbf{v}(G_i)} \mathbf{1}[P_G[v] = P_P[v]]\in \left[0, \sum_{G_i \in G} |\text{v}(G_i)|\right],
\end{equation}
where $\mathbf{v}(\cdot)$ denotes the function that maps a graph $G_i$ to its corresponding parameter value set. The normalized correctness score is then computed as:
\begin{equation}
 \mathcal{S}_{i,\text{acc}}=\frac{{r}_{\text{name}}+{r}_{\text{para}}+{r}_{\text{value}}}{1+|G|+\sum_{G_i \in G} |\text{v}(G_i)|}.   
\end{equation} 
The final dynamic scaling reward \(\mathcal{R}_{\text{final}}\) is then calculated as a weighted sum:
\begin{equation}
\mathcal{R}_{i,\text{final}}=\beta_a\mathcal{S}_{i,\text{acc}}*(1-\bar\nu)+\beta_f\mathcal{S}_{i,\text{format}}*(1-\bar\nu),
\end{equation}
where \(\beta\) denotes the relative importance and $\bar\nu \in (0,1)$ denotes the training progress.

\subsubsection{Reshaping Token-level Policy Gradients}
ResT enables a more stable and effective use of rule-based rewards during model fine-tuning. Given a prompt containing user requirements and context, a policy \( \pi_{\theta} \) generates $G$ multi-turn response sequences composed of plain text and tool invocation segments. For the $i$-th sample in a group $G$, the advantage function \(\hat{A}_i\) are then calculated:
\begin{equation}
\hat{A}_i=\frac{\mathcal{R}_{i,\text{final}}-\frac{1}{G} \sum_{i=0}^{G-1} \mathcal{R}_{i,\text{final}}}{\sqrt{\frac{1}{G} \sum_{i=0}^{G-1} (\mathcal{R}_{i,\text{final}} - \frac{1}{G} \sum_{i=0}^{G-1} \mathcal{R}_{i,\text{final}})^2}+\delta},
\end{equation}
where \(\delta > 0\) is a numerical stability constant. 

We first tokenize the response and then partition it into four weight sets: $\tilde{w}_{t,fmt}$ (format tags), $\tilde{w}_{t,name}$ (tool names), $\tilde{w}_{t,para}$ (tool parameters), and $\tilde{w}_{t,thk}$ (chain-of-thought). Reweighting factor \(\tilde{w}_t\) is initialized according to the entropy-based reweighting equation \ref{factor}.

Our proposed curriculum learning method dynamically adjusts the reweighting factor based on the training progress \(\nu\), prioritizing different token types at different stages. Tool names consistently receive a high weight, reflecting their critical importance. Initially, the focus is on structural correctness via high format-token weights. Subsequently, the focus shifts to semantic accuracy by annealing format weights while increasing parameter-token weights:
\begin{equation}
    \tilde{w}_{t,fmt}(\nu) = \max(w_{\min}, \, \tilde{w}_{t,fmt} - \alpha_f \cdot \nu), \quad \tilde{w}_{t,para}(\nu) = \min(w_{\max}, \, \tilde{w}_{t,para} + \alpha_p \cdot \nu),
\end{equation}
where $[w_{\min}, w_{\max}]$ denotes the clipping interval and \(\alpha\) denotes the magnitude of the update. To reduce policy-gradient variance, encourage step-by-step reasoning, and improve parameter accuracy, the weights for chain-of-thought and tool parameter tokens are synchronized:
\begin{equation}
    \tilde{w}_{t,thk}(\nu) = \min(w_{\max}, \, \tilde{w}_{t,thk} + \alpha_t \cdot \nu).
\end{equation}
We obtain the normalized reweighting factor by its mean over the sequence:
\begin{equation}
    \hat{w}_t := \max\left(w_{\min},\, \min\left(\tilde{w}_t,\, w_{\max}\right)\right), \quad \bar{w} := \frac{1}{|T|} \sum_{t =1}^{T} \hat{w}_t, \quad w_t = \frac{\hat{w}_t}{ \bar{w}+\delta },
\end{equation}
This normalization design enables the redistribution of learning signals among all tokens within the same sample, according to their relative policy entropy.
By integrating the advantage function with the reweighting factor, we obtain the final optimization objective:
\begin{equation}
\mathcal{L}_{\mathrm{ResT}}(\theta) = 
- \frac{1}{G} \sum_{i=0}^{G-1} \sum_{t=0}^{T-1} 
\frac{\omega_t}{T} \cdot 
\min\left( 
  r_{i,t} \hat{A}_i,\;
  \operatorname{clip}(r_{i,t}, 1-\epsilon, 1+\epsilon) \hat{A}_i 
\right),
\end{equation}
where $ r_{i,t} = \dfrac{\pi_{\theta}(y_{i,t} \mid x, y_{i,<t})}{\pi_{\theta_{\text{old}}}(y_{i,t} \mid x, y_{i,<t})} $ and $\epsilon$ limits the magnitude of gradient updates. The objective guides the policy toward structurally consistent and semantically accurate tool calls, while gradient reshaping alleviates variance in policy gradients, resulting in more stable and efficient alignment with task requirements. The complete algorithmic procedure is provided in Appendix \ref{procedure}.

\begin{table}[h]
  \centering
\caption{\textbf{BFCL multi-turn results (updated June 14, 2025).} Metrics computed with official scripts. TSFT scales loss on tool-call tokens. RSFT scales loss on reasoning tokens. Best in \textbf{bold}.}
  \label{bfcl-results}
  \resizebox{\linewidth}{!}{
  \begin{tabular}{lccccc}
 \hline
    \textbf{Model} & \textbf{Overall Acc} & \textbf{Base} & \textbf{Miss Func} & \textbf{Miss Param} & \textbf{Long Context} \\
\hline 
    Qwen3-1.7B (Base) & 8.62\% & 12.00\% & 5.50\% & 9.50\% & 7.50\% \\
    Qwen3-1.7B (SFT) & 14.25\% & 17.50\% & 14.50\% & 14.50\% & 10.50\% \\
    Qwen3-1.7B (TSFT) & 13.50\% & 18.00\% & 14.50\% &11.50\% & 10.00\% \\
    Qwen3-1.7B (RSFT) & 15.00\% & 17.00\% & 16.50\% &14.50\% &12.00\% \\
    Qwen3-1.7B (GRPO) & 14.50\% & 17.00\% & 15.50\% & 12.50\% & 13.00\% \\
    Qwen3-1.7B (SFT+GRPO) & 14.88\% & 20.50\% & 18.00\% & 9.00\% & 12.00\% \\
    Qwen3-1.7B (Ours, ResT) & \textbf{16.00\%} & 20.00\% & 16.00\% & 15.00\% & 13.00\% \\
 \hline 
    Qwen3-4B-2507 (Base) & 48.00\% & 57.00\% & 53.00\% & 33.50\% & 48.50\% \\
    Qwen3-4B-2507 (SFT) & 43.00\% & 52.50\% & 46.50\% & 33.00\% & 40.00\% \\
    Qwen3-4B-2507 (TSFT)  & 48.25\% & 58.00\% & 52.00\% & 32.50\% & 50.50\% \\
    Qwen3-4B-2507 (RSFT) & 41.25\% & 52.00\% & 47.50\% & 31.50\% & 34.00\%\\
    Qwen3-4B-2507 (GRPO) & 41.62\% & 48.50\% & 45.50\% & 27.00\% & 45.50\% \\
    Qwen3-4B-2507 (SFT+GRPO) & 40.38\% & 46.00\% & 42.00\% & 29.50\% & 44.00\% \\
    Qwen3-4B-2507 (Ours, ResT) & \textbf{50.38}\% & 62.50\% & 56.00\% & 33.00\% & 50.00\% \\    
 \hline 
    Qwen3-8B (Base) & 33.00\% & 38.50\% & 32.00\% & 31.50\% & 30.00\% \\
    Qwen3-8B (SFT) & 28.75\% & 34.50\% & 33.50\% & 22.00\% & 25.00\% \\
    Qwen3-8B (TSFT) & 29.00\% & 32.50\% & 34.00\% & 24.00\% & 25.50\% \\
    Qwen3-8B (RSFT) &25.12\% & 33.00\% & 29.50\% & 19.00\% & 19.00\% \\
    Qwen3-8B (GRPO) & 36.00\% & 42.00\% & 42.00\% & 28.50\% & 31.50\% \\
    Qwen3-8B (SFT+GRPO) & 32.38\% & 42.50\% & 35.00\% & 27.50\% & 24.50\% \\
    Qwen3-8B (Ours, ResT) & \textbf{40.13\%} & 50.50\% & 45.00\% & 32.00\% & 33.00\% \\

\hline  
    Qwen3-14B (Base) & 36.62\% & 48.00\% & 33.00\% & 33.50\% & 32.00\% \\
    Qwen3-14B (SFT) & 28.12\% & 33.50\% & 30.00\% & 23.00\% & 26.00\% \\
    Qwen3-14B (TSFT) & 27.62\% & 34.50\% & 34.00\% & 17.50\% & 24.50\% \\
    Qwen3-14B (RSFT) & 27.00\% & 33.50\% & 27.50\% & 20.00\% & 27.00\% \\
    Qwen3-14B (GRPO) & 38.88\% & 50.00\% & 43.50\% & 26.50\% & 35.50\% \\
    Qwen3-14B (SFT+GRPO) & 34.12\% & 45.00\% & 37.50\% & 22.00\% & 32.00\% \\
    Qwen3-14B (Ours, ResT) & \textbf{44.25\%} & 57.00\% & 45.00\% & 36.50\% & 38.50\% \\

 \hline 
GPT-4o-2024-11-20 & {50.00\%} & 61.00\% & 45.50\% & 35.50\% & 58.00\% \\
GPT-4.5-Preview-2025-02-27  & 45.38\% & 58.00\% & 37.50\% & 43.00\% & 43.00\% \\
GPT-4.1-2025-04-14 & 40.50\% & 48.00\% & 34.00\% & 35.00\% & 45.00\% \\
o3-mini-2025-01-31 & 28.75\% & 32.00\% & 26.00\% & 23.00\% & 34.00\% \\
Grok-3-beta & 45.50\% &	54.00\% & 39.00\% & 42.00\% & 47.00\% \\
DeepSeek-V3-0324 & 29.87\% & 41.00\% & 21.00\% & 23.00\% & 34.50\% \\
DeepSeek-R1-0528 & 38.88\% & 49.50\% & 31.50\% & 32.00\% & 42.50\% \\
Gemini-2.5-Pro-Preview-05-06 & 34.62\% & 39.50\% & 29.50\% & 31.50\% & 38.00\% \\
 \hline 
  \end{tabular}}
  \vspace{-1.0em}
\end{table}

\section{Experiments}
\label{experiments}
\subsection{Training Details}
We curate a 4k mixed corpus to facilitate robust tool learning in RL: ToolACE \citep{liu2025toolace} (2k samples) targets the invoke-vs-answer decision in multi-step interactions; Hammer (masked) \citep{lin2025robusthammer} (1k samples) randomizes tool and argument names to enforce description-grounded generalization; and XLAM \citep{zhang2024xlamfamilylargeaction} (1k samples) supplies compositional tasks that require one or multiple tool calls per turn. Together, these sources supervise invocation timing, robust grounding, and compositional planning \citep{qian2025toolrl}.

We convert multi-turn dialogue samples into single-step instances to provide denser and more fine-grained reward signals. Concretely, a dialogue with $K$ interaction steps is decomposed into $K$ sub-instances, each conditioned on the full preceding conversation as input and supervised with the model’s current action (e.g., tool invocation or response) as the target. This decomposition strategy, adopted from the SWiRL framework \citep{goldie2025syntheticdatageneration}, substantially increases the density of training signals. Consequently, the model receives process-level supervision at each step, rather than only for the final outcome. This step-wise supervision fosters more stable learning of multi-step reasoning and tool use.

\begin{table}[h]
  \centering
  \caption{\textbf{API-Bank Test Results.} The results presented correspond to the highest score achieved by each method with its optimal hyperparameter settings. The evaluation dataset consists of 399 samples for Level 1, 67 for Level 2, and 131 for Level 3. Best in \textbf{bold}.}
  \label{api-results}
  \begin{tabular}{lcccc}
\hline 
    \textbf{Model} & \textbf{Overall Acc} & \textbf{Level 1 Acc} & \textbf{Level 2 Acc} & \textbf{Level 3 Acc} \\
     \hline
    Qwen3-1.7B (Base) & 47.57\% & 53.38\% & 28.36\% & 39.69\% \\
    Qwen3-1.7B (SFT)& 58.96\% & 66.92\% & 52.24\% & 38.17\% \\
    Qwen3-1.7B (TSFT) & 54.77\% & 61.4\% & 43.28\% & 40.46\% \\
    Qwen3-1.7B (RSFT) & 61.81\% & 67.92\% & 59.7\% & 44.27\% \\
    Qwen3-1.7B (GRPO) & 63.65\% & 70.68\% & 61.19\% & 41.22\% \\
    Qwen3-1.7B (SFT+GRPO) & 59.46\% & 65.91\% & 58.21\% & 40.46\% \\
    Qwen3-1.7B (Ours, ResT) & \textbf{64.99\%} & 70.18\% & 65.67\% & 48.85\% \\

 \hline
    Qwen3-4B-2507 (Base) & 66.33\% & 72.68\% & 64.18\% & 48.09\% \\
    Qwen3-4B-2507 (SFT)& 60.30\% & 67.92\% & 59.70\% & 37.40\% \\
    Qwen3-4B-2507 (TSFT)& 66.83\% & 72.93\% & 64.18\% & 49.62\% \\
    Qwen3-4B-2507 (RSFT) & 64.49\% & 69.92\% & 59.7\% & 50.38\% \\
    Qwen3-4B-2507 (GRPO) & 65.33\% & 71.93\% & 62.69\% & 46.56\% \\
    Qwen3-4B-2507 (SFT+GRPO) & 68.51\% & 74.44\% & 65.67\% & 51.91\% \\
    Qwen3-4B-2507 (Ours, ResT) & \textbf{68.68\%} & 73.68\% & 64.18\% & 55.73\%     \\

 \hline 
    Qwen3-8B (Base) & 63.32\% & 70.68\% & 53.73\% & 45.80\% \\
    Qwen3-8B (SFT)& 60.64\% & 69.17\% & 56.72\% & 36.64\% \\
    Qwen3-8B (TSFT) & 59.80\% & 66.42\% & 50.75\% & 44.27\% \\
    Qwen3-8B (RSFT) & 60.47\% & 67.42\% & 53.73\% & 42.75\% \\
    Qwen3-8B (GRPO) & 66.50\% & 75.44\% & 64.18\% & 40.46\% \\
    Qwen3-8B (SFT+GRPO) & 63.82\% & 72.18\% & 52.22\% & 42.75\% \\
    Qwen3-8B (Ours, ResT) & \textbf{70.69\%} & 74.94\% & 65.67\% & 60.31\% \\
 \hline
    Qwen3-14B (Base) & 65.49\% & 71.68\% & 56.72\% & 51.15\% \\
    Qwen3-14B (SFT) & 67.34\% & 76.44\% & 62.69\% & 41.98\% \\
    Qwen3-14B (TSFT) & 64.32\% & 68.67\% & 59.70\% & 53.44\% \\
    Qwen3-14B (RSFT) & 64.99\% & 70.68\% & 55.22\% & 52.67\% \\
    Qwen3-14B (GRPO) & 66.33\% & 73.18\% & 59.70\% & 48.85\% \\
    Qwen3-14B (SFT+GRPO) & 67.34\% & 73.43\% & 58.21\% & 53.44\% \\
    Qwen3-14B (Ours, ResT) & \textbf{69.35\%} & 73.43\% & 62.69\% & 60.31\% \\
 \hline 
  \end{tabular}
\vspace{-2.0em}
\end{table}

\begin{figure}[h]
\centering
\vspace{-1.0em}
\label{curves}
\setlength{\fboxsep}{0pt}
\includegraphics[height=4.5cm, width=0.95\linewidth]{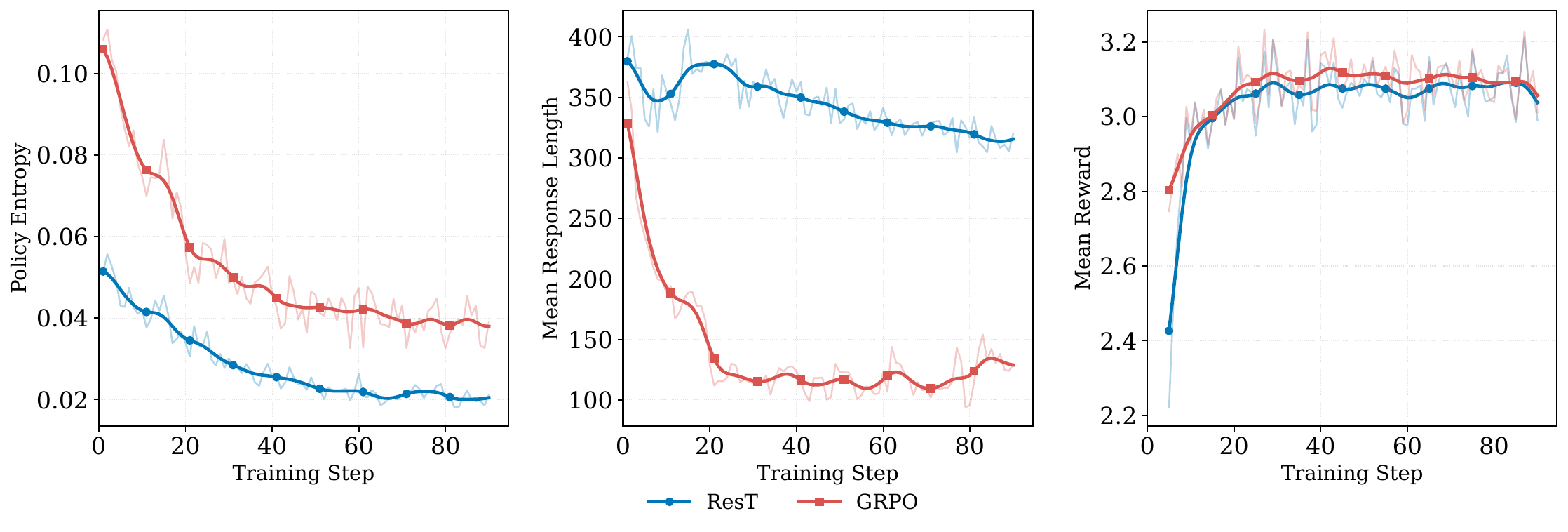}
\caption{Learning curves for ResT and GRPO during training steps. The training dynamics show that ResT achieves a significantly lower and smoother policy entropy compared to GRPO, while maintaining comparable reward performance and longer responses.}
\label{fig:curves}
\vspace{-1.0em}
\end{figure}

\subsection{Main Results}
We evaluate six baselines spanning supervised and reinforcement-learning regimes:
\textbf{Base Model}: instruction-tuned without task-specific fine-tuning.
\textbf{SFT}: supervised fine-tuning on the task training set.
\textbf{TSFT}: SFT with up-weighted loss \citep{huerta-enochian-ko-2024-instruction} on tool-call tokens to prioritize invocation accuracy.
\textbf{RSFT}: SFT with up-weighted loss on reasoning tokens to strengthen intermediate reasoning.
\textbf{GRPO}: RL-only method representative of the current state of the art for tool use.
\textbf{SFT+GRPO}: initialize from an SFT checkpoint, then apply GRPO to isolate RL’s marginal gains.

\textbf{Benchmarks and Setup.} The proposed method is rigorously evaluated on two comprehensive benchmarks: the Berkeley Function Calling Leaderboard (BFCL) \citep{patil2023bfcl}, which assesses capabilities like multi-step tool use, and API-Bank \citep{li2023apibankcomprehensivebenchmarktoolaugmented}, a three-tiered suite testing tool invocation in multi-turn dialogues. To isolate the algorithm's contribution, several open-source models are fine-tuned against strong domain-specific baselines.

\textbf{Performance Results.} As reported in Figure \ref{fig:result}, Table \ref{bfcl-results} and Table \ref{api-results}, fine-tuning the Qwen3 model family with ResT establishes a new state-of-the-art on both benchmarks. The method yields significant performance gains over the GRPO baseline, with improvements of up to $8.76\%$ on BFCL and $3.02\%$ on API-Bank under an identical data budget. Notably, the Qwen3-4B-2507 model fine-tuned with ResT even surpasses GPT-4o by $0.38\%$ on multi-turn overall tool-use tasks, especially by $1.50\%$ on multi-turn base tasks. On the single-turn tool-use tasks, it leads by $4.11\%$. These behaviors indicate emergent forms of proactivity and metacognition, which enhance efficiency, mitigate hallucinations, and signal the emergence of foundational agentic intelligence.

\textbf{Training Dynamics Analysis.} The learning curves in Figure \ref{fig:curves} highlight ResT’s advantage over GRPO. ResT’s policy entropy converges to \(\approx 0.02\)—lower and less volatile than GRPO’s \(\approx 0.04\)—indicating a more deterministic policy. This stability does not reduce performance: both reach similar final mean reward. Behaviorally, ResT sustains longer outputs (mean \(>300\) tokens), whereas GRPO collapses to \(\approx 125\) tokens. Overall, ResT stabilizes optimization and yields longer, more complex responses without sacrificing reward, avoiding the degradation observed with GRPO.

\begin{table}[h]
  \centering
\vspace{-1.0em}
  \caption{\textbf{API-Bank Test Results for Relaxations.} No dynamic reward denotes no dynamic scaling reward value. No gradients for CoT denotes no gradient update for chain-of-thought. No curriculum learning assigns a constant weight to each segment. Best in \textbf{bold}.}
  \label{ablation}
   \resizebox{\linewidth}{!}{
  \begin{tabular}{lcccc}
\hline 
    \textbf{Model} & \textbf{Overall Acc} & \textbf{Level 1 Acc} & \textbf{Level 2 Acc} & \textbf{Level 3 Acc} \\
 \hline 
Qwen3-1.7B (\textit{No dynamic reward}) & 64.15\% & 69.92\% & 58.21\% & 49.62\% \\
Qwen3-1.7B (\textit{No gradients for CoT}) & 62.98\% & 68.92\% & 62.69\% & 45.04\% \\
Qwen3-1.7B (\textit{No curriculum learning}) & {61.14\%} & {67.92\%} & 58.21\% & {41.98\%} \\
Qwen3-1.7B (Ours, ResT) & {\textbf{64.99\%}} & {70.18\%} & 65.67\% & {48.85\%} \\    
 \hline
Qwen3-4B-2507 (\textit{No dynamic reward}) & 64.15\% & 71.18\% & 64.18\% & 42.75\% \\
Qwen3-4B-2507 (\textit{No gradients for CoT}) & {64.49\%} & 71.18\% & {59.70\%} & {46.56\%} \\
Qwen3-4B-2507 (\textit{No curriculum learning}) & {67.84\%} & {73.18\%} & 64.18\% & {53.44\%} \\
Qwen3-4B-2507 (Ours, ResT) & {\textbf{68.68\%}} & {73.68\%} & 64.18\% & {55.73\%} \\    
 \hline
    
Qwen3-8B (\textit{No dynamic reward}) & 64.15\% & 71.68\% & 64.18\% & 41.22\% \\
Qwen3-8B (\textit{No gradients for CoT}) & {66.33\%} & {74.44\%} & {65.67\%} & {41.98\%} \\
Qwen3-8B (\textit{No curriculum learning}) & {65.83\%} & {73.68\%} & {62.69\%} & {43.51\%} \\
Qwen3-8B (Ours, ResT) & {\textbf{70.69\%}} & {74.94\%} & {65.67\%} & {60.31\%} \\

 \hline 
  \end{tabular}}
  \vspace{-1.0em}
\end{table}

\subsection{Ablation Study}
Ablations across base models (Table~\ref{ablation}) show that each ResT component is necessary for peak performance. Disabling dynamic rewards, CoT policy gradients, or the curriculum consistently reduces accuracy. On Qwen3-8B, removing dynamic rewards lowers accuracy by 6.54\%, and dropping CoT policy gradients and the curriculum yields declines of 4.36\% and 4.86\%, respectively. These results indicate complementary contributions and explain ResT’s robustness.

The ResT-trained model demonstrates robust generalization beyond its multi-turn training focus. As shown in Table \ref{bfcl-ablation}, it performs comparably to the baseline on single-turn tasks and exhibits superior performance on real-time updated tasks, underscoring its out-of-distribution generalization. We also extend our evaluation to the Llama-3.2-3B-Instruct fine-tuned with ResT in Table \ref{Llama} and Table \ref{bfcl-Llama} to verify that the effectiveness of our method is not confined to a specific model family. This confirms that the benefits of our approach are model-agnostic and highlights its broad applicability.
\begin{table}[h]
  \centering
  \vspace{-1.0em}
  \caption{\textbf{BFCL Results on Llama3.2-3B-Instruct model.} All metrics are calculated using the official scripts. Overall accuracies for the three different tasks are provided separately. Best in \textbf{bold}.}
  \label{bfcl-Llama}
 \resizebox{\linewidth}{!}{
  \begin{tabular}{lccccc}
\\ \hline 
    \textbf{Model} & \textbf{Non-Live AST Acc} & \textbf{Live Acc} &\textbf{Multi Turn Acc} & \textbf{Relevance Detection} & \textbf{Irrelevance Detection} \\
\hline 
Llama-3.2-3B-Instruct (Base) & 17.44\%  & 43.85\% & 0.00\% & 77.78\%& 66.07\% \\ 
Llama-3.2-3B-Instruct (SFT) &  65.42\%  & 63.04\% & 1.38\% & 77.78\% & 78.25\% \\
Llama-3.2-3B-Instruct (TSFT) &  57.06\%  & 45.45\% & 0.12\% & 38.89\% & 59.17\% \\
Llama-3.2-3B-Instruct (RSFT) &  35.35\%  & 60.20\% & 1.25\% & 94.44\% & 71.00\% \\
Llama-3.2-3B-Instruct (GRPO) & {74.38\%} & {56.86\%} & {1.37\%} & {94.44\%} & 62.23\% \\ 
Llama-3.2-3B-Instruct (SFT+GRPO) &  23.92\%  & 45.40\% & 0.00\% & 11.11\% & 89.99\% \\
Llama-3.2-3B-Instruct (Ours, ResT) & \textbf{83.19\%} & \textbf{61.53\%} & \textbf{3.50\%} & {88.89\%} & 66.56\% \\ 
 \hline 
  \end{tabular}}
  \vspace{-1.0em}
  \end{table}
\section{Conclusion}
We study tool-use in LLM agents through the lens of RL and establish a theoretical link between token-level policy entropy and training stability: lowering entropy tightens variance bounds on policy gradients, and reward mass concentrates on structured, low-entropy tokens such as tool names, arguments, and output-format markers. Building on this insight, we propose ResT, an entropy-aware, token-level RL algorithm that reshapes policy gradients via entropy-informed reweighting and a lightweight curriculum that progressively upweights reasoning tokens. This enables a smooth shift from structural correctness to semantic reasoning and stabilizes convergence in long-horizon, multi-turn tool use. Empirically, ResT attains state-of-the-art results on BFCL and API-Bank, improving prior methods by up to $8.76\%$; when fine-tuned on a 4B base LLM, it further surpasses GPT-4o by $4.11\%$ on single-turn tasks and $1.50\%$ on multi-turn base tasks. Future work will integrate verifiable structural rewards with judge-based semantic scores, learn schedules for entropy-aware mixing, and calibrate LLM-as-a-judge with uncertainty estimates.

\clearpage
\bibliography{references}  
\clearpage
\appendix
\section{Proofs}
\label{appendix:proofs}

This appendix provides detailed proofs of the theoretical results presented in Section \ref{theory}.

\subsection{Proof of Lemma~\ref{lem:var-decomp} (Policy Gradient Variance Decomposition)}
\label{appendix:proof_lemma_var_decomp}
\begin{proof}
\textbf{Setup.}
Let a trajectory be $\tau=(x,y_{0:T-1})$ sampled from $\pi_\theta$, with return $R(\tau)$.
For trajectory index $i$, write $\tau_i$, $r_i:=R(\tau_i)$, and $y_{i,t}$.
The per-time-step score is $\nabla_\theta\log\pi_\theta(y_{i,t}\mid x,y_{i,<t})$.

\textbf{Step 1. Policy gradient with a baseline.}
By the policy gradient theorem,
\[
\nabla_\theta J(\theta)
=\E_{\tau\sim\pi_\theta}\!\Biggl[\sum_{t=0}^{T-1}
   \nabla_\theta\log\pi_\theta(y_t\mid x,y_{<t})\,R(\tau)\Biggr].
\]
Let $b_t=b_t(x,y_{<t})$ be any baseline that does not depend on the sampled action $y_t$ given the state/history.
Then, conditioning on $(x,y_{<t})$,
\[
\begin{aligned}
\E_{y_t\sim\pi_\theta}\!\big[\nabla_\theta\log\pi_\theta(y_t\mid x,y_{<t})\,b_t\big]
&= b_t\,\sum_{y_t}\pi_\theta(y_t\mid x,y_{<t})\,\nabla_\theta\log\pi_\theta(y_t\mid x,y_{<t})\\
&= b_t\,\nabla_\theta\!\sum_{y_t}\pi_\theta(y_t\mid x,y_{<t}) \;=\; 0.
\end{aligned}
\]

Hence one may replace $R(\tau)$ by the advantage $A(\tau):=R(\tau)-\sum_t b_t$ inside the expectation without changing its value. In what follows we set $b_t\equiv 0$ for simplicity (equivalently, redefine $\mu,\sigma$ as the population moments of $A(\tau)$ if a baseline is used).

\textbf{Step 2. Population standardization and unbiased direction.}
Consider the population-standardized advantage
\[
\hat A_i \;=\; \frac{r_i-\mu}{\sigma},\qquad \mu:=\E[r],\ \ \sigma^2:=\Var(r),
\]
where expectations are under the current policy $\pi_\theta$.
Define the trajectory-level gradient
\[
g_i \;=\; \hat A_i \sum_{t=0}^{T-1}
\nabla_\theta\log\pi_\theta(y_{i,t}\mid x,y_{i,<t}).
\]
Taking expectations and using $\E[r_i-\mu]=0$,
\[
\E[g_i]
=\frac{1}{\sigma}\,\E\!\Biggl[(r_i-\mu)\sum_{t}\nabla_\theta\log\pi_\theta(y_{i,t}\mid x,y_{i,<t})\Biggr]
=\frac{1}{\sigma}\,\nabla_\theta J(\theta).
\]
Thus $g_i$ is an unbiased \emph{direction} estimator for $\nabla_\theta J(\theta)$, up to a scalar factor $1/\sigma$.
If instead one uses the centered (unnormalized) advantage $\hat A_i=r_i-\mu$, then $\E[g_i]=\nabla_\theta J(\theta)$ exactly.
In practice the factor $1/\sigma$ can be absorbed into the learning rate.

\textbf{Step 3. Mini-batch variance decomposition.}
Let $g_1,\dots,g_G$ be i.i.d.\ copies of $g_i$ with finite second moments, and define the mini-batch estimator
\[
\bar g \;=\; \frac{1}{G}\sum_{i=1}^G g_i.
\]
Its covariance is
\[
\Cov(\bar g)
=\E\!\big[(\bar g-\E[g_1])(\bar g-\E[g_1])^\top\big]
=\frac{1}{G^2}\sum_{i=1}^G\sum_{j=1}^G \Cov(g_i,g_j).
\]
Independence gives $\Cov(g_i,g_j)=0$ for $i\neq j$, hence
\[
\Cov(\bar g)=\frac{1}{G^2}\sum_{i=1}^G \Cov(g_i,g_i)
=\frac{1}{G}\,\Cov(g_1).
\]
Taking the $k$-th diagonal entry yields, for any coordinate $k$,
\[
\Var\!\big(\bar g^{(k)}\big)=\frac{1}{G}\,\Var\!\big(g_1^{(k)}\big),
\]
proving \eqref{eq:var-mean}.

\textbf{Step 4. Consistency checks.}
All symbols are trajectory-indexed ($i$) or time-indexed ($t$) but not conflated.
The expectation and variance are with respect to the data-generating policy $\pi_\theta$.
Population moments $(\mu,\sigma)$ refer to returns under $\pi_\theta$ at the current parameters.
The i.i.d.\ assumption holds when trajectories are drawn independently from $\pi_\theta$.
\end{proof}

\noindent\emph{Remark (batch normalization of advantages).}
If one instead uses the \emph{batch-normalized} statistic
\[
\hat A_i \;=\; \frac{r_i-\mu_{\mathcal G}}{\sigma_{\mathcal G}},\qquad
\mu_{\mathcal G}:=\frac{1}{G}\sum_{j=1}^G r_j,\ \ 
\sigma_{\mathcal G}^2:=\frac{1}{G}\sum_{j=1}^G (r_j-\mu_{\mathcal G})^2,
\]
then $\{g_i\}$ are no longer independent because each $g_i$ is a function of the full batch.
A direct expansion gives the exact decomposition
\[
\Cov(\bar g)
=\frac{1}{G}\Var(g_1)+\frac{G-1}{G}\,\Cov(g_1,g_2),
\]
where $\Cov(g_1,g_2)$ denotes the cross-covariance induced solely by the shared batch statistics.
Under mild moment and smoothness conditions, a delta-method (influence function) expansion of $(\mu_{\mathcal G},\sigma_{\mathcal G})$ around $(\mu,\sigma)$ yields $\Cov(g_1,g_2)=O(1/G)$, so
\[
\Cov(\bar g)=\frac{1}{G}\Var(g_1)+O\!\left(\frac{1}{G}\right),
\]
and the bias of $\E[\bar g]$ relative to $\nabla_\theta J(\theta)/\sigma$ is $O(1/G)$.
Both corrections vanish as $G\to\infty$, but for small $G$ they can be non-negligible in practice.

\clearpage

\subsection{Proof of Lemma~\ref{prop:moment-entropy} (Second-Order Moment–Entropy Connection)}
\label{appendix:proof_prop_moment_entropy}
\begin{proof}
\textbf{Setup and notation.}
Fix a single trajectory. Let $z_t\in\mathbb{R}^V$ denote the logits, $p_t=\mathrm{softmax}(z_t)\in\Delta^{V-1}$ the token distribution, and $e_{y_t}$ the one-hot vector of $y_t\sim p_t$. Define the score vector $s_t:=e_{y_t}-p_t\in\mathbb{R}^V$ and the Jacobian $J_t:=\partial z_t/\partial\theta\in\mathbb{R}^{V\times d}$. Then
\[
\nabla_\theta \log \pi_\theta(y_t\mid x,y_{<t}) \;=\; J_t^\top s_t,
\quad
g_i \;=\; \hat A_i \sum_{t=0}^{T-1} J_t^\top s_t .
\]
All expectations below are taken over the joint randomness of the trajectory and the policy unless explicitly conditioned.

\textbf{Step 1. Second-order moment control.}
By the definition of vector variance,
\[
\Var(g_i) \;=\; \E\!\left[\left\|g_i\right\|^2\right]-\left\|\E[g_i]\right\|^2
\;\le\; \E\!\left[\left\|g_i\right\|^2\right]
\;=\; \E\!\Biggl[\hat A_i^{\,2}\Biggl\|\sum_{t=0}^{T-1}J_t^\top s_t\Biggr\|^2\Biggr].
\tag{$\ast$}\label{eq:moment-upper}
\]

Condition on the trajectory prefix up to time $t$. Since 
\[
\E[s_t \mid x,y_{<t}] \;=\; 0,
\]
the sequence $\{J_t^\top s_t\}_{t=0}^{T-1}$ forms a vector-valued martingale difference, and thus is $L^2$-orthogonal. Therefore
\[
\E\!\Biggl[\Biggl\|\sum_{t=0}^{T-1}J_t^\top s_t\Biggr\|^2\Biggr]
=\sum_{t=0}^{T-1}\E\!\left[\|J_t^\top s_t\|^2\right].
\]

Applying the inequality $\|J_t^\top s_t\|\le \|J_t\|_F\|s_t\|$ gives
\[
\E\!\Biggl[\hat A_i^{\,2}\Biggl\|\sum_{t=0}^{T-1}J_t^\top s_t\Biggr\|^2\Biggr]
\;\le\; \E[\hat A_i^{\,2}] \sum_{t=0}^{T-1} \E\!\big[\|J_t\|_F^2\,\|s_t\|^2\big]
\;=\; \E[\hat A_i^{\,2}] \sum_{t=0}^{T-1} \beta_t\,\E\!\big[\|s_t\|^2\big],
\]
where we set
\[
\beta_t \;:=\; \E\!\big[\|J_t\|_F^2\big] \quad\text{or more tightly}\quad
\beta_t \;:=\; \E\!\big[\|J_t\|_F^2(1-e^{-H_t})\big] \;\ge 0,
\]
depending on whether we insert the entropy factor at this stage. Either choice preserves the quadratic-in-weights structure used later; we adopt the tighter definition when convenient.

\textbf{Step 2. Exact expression for the expected score norm.}
Condition on $p_t$. Since $s_t=e_{y_t}-p_t$ with $y_t\sim p_t$,
\[
\|s_t\|^2 \;=\; \|e_{y_t}\|^2 -2\langle e_{y_t},p_t\rangle + \|p_t\|^2
\;=\; 1 - 2p_{t,y_t} + \sum_{v=1}^V p_{t,v}^2 .
\]
Taking expectation over $y_t\sim p_t$ removes the index:
\[
\E_{y_t\sim p_t}[\|s_t\|^2 \mid p_t]
\;=\; 1 - \sum_{v=1}^V p_{t,v}^2
\;=\; 1 - \|p_t\|_2^2.
\tag{$\dagger$}\label{eq:score-l2}
\]

\textbf{Step 3. Entropy connection via the ordering $H_2 \le H$.}
Since the Rényi-2 entropy $H_2(p_t)=-\log\sum_v p_{t,v}^2$ is always upper-bounded by the Shannon entropy $H_t$, we obtain
\[
H_t \;=\; \E_{v\sim p_t}\!\big[-\log p_{t,v}\big] \;\ge\; -\log \E_{v\sim p_t}\![p_{t,v}]
\;=\; -\log\!\Big(\sum_{v=1}^V p_{t,v}^2\Big).
\]
Equivalently,
\[
\sum_{v=1}^V p_{t,v}^2 \;\ge\; e^{-H_t}.
\tag{$\ddagger$}\label{eq:renyi2}
\]
Combining \eqref{eq:score-l2} and \eqref{eq:renyi2} gives the entropy upper bound
\[
\E[\|s_t\|^2 \mid p_t] \;=\; 1-\|p_t\|_2^2 \;\le\; 1-e^{-H_t}.
\]
Taking expectation over the randomness in $p_t$ yields
\[
\E[\|s_t\|^2] \;\le\; 1-\E[e^{-H_t}]
\;\le\; 1-e^{-\E[H_t]},
\]
where the last inequality uses convexity of $\exp(\cdot)$ (optional if one keeps $1-\E[e^{-H_t}]$). Equality in \eqref{eq:renyi2} holds iff $p_t$ is uniform on its support, certifying tightness.

\textbf{Step 4. Conclusion.}
Substituting the entropy bound for $\E[\|s_t\|^2]$ into \eqref{eq:moment-upper} gives
\[
\Var(g_i)
\;\le\; \E[\hat A_i^{\,2}] \sum_{t=0}^{T-1} \E\!\big[\|J_t\|_F^2\,\|s_t\|^2\big]
\;\le\; \E[\hat A_i^{\,2}] \sum_{t=0}^{T-1} \E\!\big[\|J_t\|_F^2(1-e^{-H_t})\big],
\]
which is precisely the claimed second-order moment–entropy relation.
\medskip

\noindent\emph{Remarks on assumptions and tightness.}
(i) The step factoring $\E[\hat A_i^{\,2}]$ is exact if $\hat A_i$ is independent of $(J_t,s_t)$ or after conditioning on the trajectory prefix; without independence one may apply Hölder to absorb the mismatch into a constant, leaving the functional dependence on entropy unchanged.
(ii) The bound $\E[\|s_t\|^2]=1-\|p_t\|_2^2\le 1-e^{-H_t}$ becomes tight when $p_t$ is uniform over its support, reflecting the ordering $H_2\le H$ between R\'enyi-2 and Shannon entropies.
\end{proof}

\clearpage
\subsection{Proof of Theorem~\ref{thm:opt-weights} (Optimal Entropy-Aware Reweighting)}
\label{appendix:proof_main}
\begin{proof}
\textbf{Step 1. Starting point.}
By Lemma~\ref{prop:moment-entropy} and the inequality $\|J_t^\top s_t\|^2 \le \|J_t\|_F^2\|s_t\|^2$,
\[
\Var(g_i)\;\le\; \E\!\Bigl[\hat A_i^{\,2}\sum_{t=0}^{T-1}\|J_t\|_F^2\,\|s_t\|^2\Bigr].
\]
Applying the entropy bound $\E[\|s_t\|^2]\le 1-e^{-H_t}$ gives
\[
\Var(g_i)\;\le\; \E\!\bigl[\hat A_i^{\,2}\bigr]\sum_{t=0}^{T-1}\E\!\bigl[\|J_t\|_F^2(1-e^{-H_t})\bigr]
\;=\; \E\!\bigl[\hat A_i^{\,2}\bigr]\sum_{t=0}^{T-1}\beta_t,
\]
where we set
\[
\beta_t\;:=\;\E\!\bigl[\|J_t\|_F^2(1-e^{-H_t})\bigr]\ \ge\ 0.
\]
\emph{Remark.} Factoring out $\E[\hat A_i^{\,2}]$ tacitly assumes that $\hat A_i$ is independent of $(J_t,H_t)$ or that the bound is taken conditionally on the trajectory; without independence one may invoke Hölder/Cauchy--Schwarz to obtain an equivalent quadratic objective in the weights, which preserves the optimal proportionality derived below.

\textbf{Step 2. Effect of reweighting.}
Introduce per-step nonnegative weights $\tilde{w}_t\ge 0$ with normalization $\sum_{t=0}^{T-1}\tilde{w}_t=T$, and consider the reweighted estimator
\[
g_i^{(\mathrm{rw})}\;=\;\hat A_i\sum_{t=0}^{T-1}\tilde{w}_t\,J_t^\top s_t.
\]
Variance scales quadratically in the weights, hence
\[
\Var\!\bigl(g_i^{(\mathrm{rw})}\bigr)
\;\le\;
\E\!\bigl[\hat A_i^{\,2}\bigr]\sum_{t=0}^{T-1}\beta_t\,(\tilde{w}_t)^{2}.
\]

\textbf{Step 3. Convex program.}
We minimize the separable quadratic
\[
\min_{\{\tilde{w}_t\}}\ \sum_{t=0}^{T-1}\beta_t\,(\tilde{w}_t)^{2}
\quad\text{s.t.}\quad
\sum_{t=0}^{T-1}\tilde{w}_t=T,\quad \tilde{w}_t\ge 0.
\]
If $\beta_t>0$ for all $t$, the objective is strictly convex and the feasible set is a simplex, so the minimizer is unique.

\textbf{Step 4. Lagrangian and stationarity.}
We first solve the equality-constrained problem (the inequality will be verified inactive). The Lagrangian reads
\[
\mathcal{L}(\tilde w,\lambda)
=\sum_{t=0}^{T-1}\beta_t\,(\tilde{w}_t)^{2}
+\lambda\Bigl(\sum_{t=0}^{T-1}\tilde{w}_t-T\Bigr).
\]
Taking the derivative with respect to $\tilde{w}_t$ gives the stationarity condition
\[
\frac{\partial \mathcal{L}}{\partial \tilde{w}_t}
=2\beta_t\,\tilde{w}_t+\lambda=0
\quad\Longrightarrow\quad
\tilde{w}_t=-\frac{\lambda}{2\beta_t}
=\frac{c}{\beta_t}\quad\text{with }c:=-\lambda/2.
\]

\textbf{Step 5. Normalization and interiority.}
Enforcing $\sum_{t=0}^{T-1}\tilde{w}_t=T$ yields
\[
c\sum_{t=0}^{T-1}\beta_t^{-1}=T
\quad\Longrightarrow\quad
c=\frac{T}{\sum_{u=0}^{T-1}\beta_u^{-1}},\qquad
\lambda^*=-\frac{2T}{\sum_{u=0}^{T-1}\beta_u^{-1}}<0.
\]
Hence, for all $t$,
\[
\tilde{w}_t^{\star}
=\frac{T\,\beta_t^{-1}}{\sum_{u=0}^{T-1}\beta_u^{-1}}
>0,
\]
which a posteriori verifies that the nonnegativity constraints are inactive at the optimum (the KKT multipliers for the inequalities vanish). Strict convexity then guarantees uniqueness.

\textbf{Step 6. Degenerate case.}
If some $\beta_{t_0}=0$, any feasible allocation supported on $\{t:\beta_t=0\}$ attains objective value $0$, which is minimal. Otherwise, when all $\beta_t>0$, the interior solution above is the unique minimizer.

\textbf{Step 7. Optimal value and variance bound.}
Substituting $\tilde{w}^{\star}$ into the quadratic gives
\[
\sum_{t=0}^{T-1}\beta_t\,(\tilde{w}_t^{\star})^{2}
=\sum_{t=0}^{T-1}\beta_t\left(\frac{T\,\beta_t^{-1}}{\sum_{u=0}^{T-1}\beta_u^{-1}}\right)^{\!2}
=\frac{T^{2}}{\sum_{u=0}^{T-1}\beta_u^{-1}}.
\]
Therefore,
\[
\Var\!\bigl(g_i^{(\mathrm{rw})}\bigr)
\;\le\;
\E\!\bigl[\hat A_i^{\,2}\bigr]\cdot
\frac{T^{2}}{\sum_{t=0}^{T-1}\beta_t^{-1}}.
\]

\textbf{Step 8. Dual certificate via Cauchy--Schwarz.}
For any feasible $\{\tilde{w}_t\}$,
\[
\Bigl(\sum_{t=0}^{T-1}\tilde{w}_t\Bigr)^{2}
=\Bigl(\sum_{t=0}^{T-1}\sqrt{\beta_t}\,\tilde{w}_t\cdot \beta_t^{-1/2}\Bigr)^{2}
\le \Bigl(\sum_{t=0}^{T-1}\beta_t\,(\tilde{w}_t)^{2}\Bigr)
\Bigl(\sum_{t=0}^{T-1}\beta_t^{-1}\Bigr),
\]
which implies
\[
\sum_{t=0}^{T-1}\beta_t\,(\tilde{w}_t)^{2}
\ \ge\ \frac{T^{2}}{\sum_{t=0}^{T-1}\beta_t^{-1}},
\]
with equality if and only if $\tilde{w}_t\propto \beta_t^{-1}$, matching $\tilde{w}_t^{\star}$ and certifying optimality.
\end{proof}

\clearpage

\section{Training Algorithm: ResT with Rule-Based Rewards}
\label{procedure}
\begin{algorithm}[h]
\caption{Rule-Based Reward Computation}
\label{alg:reward}
\DontPrintSemicolon
\KwIn{Prediction $(N_P,P_P,\mathbf{v}_P)$; ground truth $(N_G,P_G,\mathbf{v}_G)$ for each tool; format template $\mathcal{T}$; training progress $\bar\nu\in(0,1)$; weights $(\beta_a,\beta_f)$.}
\KwOut{$\mathcal{R}_{i,\text{final}}$.}
\BlankLine
\textbf{Format score}:\;
$\mathcal{S}_{i,\text{format}} \leftarrow \mathbf{1}\!\left[\text{prediction matches }\mathcal{T}\text{ exactly}\right]$\;
\textbf{Name score}:\;
$r_{\text{name}} \leftarrow \frac{|N_G\cap N_P|}{|N_G\cup N_P|}$\;
\textbf{Parameter-name score}:\;
$r_{\text{para}} \leftarrow \sum_{G_j\in G}\frac{|P_G(G_j)\cap P_P(G_j)|}{|P_G(G_j)\cup P_P(G_j)|}$\;
\textbf{Parameter-value score}:\;
$r_{\text{value}} \leftarrow \sum_{G_j\in G}\sum_{v\in \mathbf{v}_G(G_j)} \mathbf{1}\!\left[P_G(G_j)[v]=P_P(G_j)[v]\right]$\;
\textbf{Normalization}:\;
$Z \leftarrow 1+|G|+\sum_{G_j\in G}|\mathbf{v}_G(G_j)|$\;
$\mathcal{S}_{i,\text{acc}} \leftarrow \dfrac{r_{\text{name}}+r_{\text{para}}+r_{\text{value}}}{Z}$\;
\textbf{Dynamic scaling}:\;
$\mathcal{R}_{i,\text{final}} \leftarrow (1-\bar\nu)\left(\beta_a\,\mathcal{S}_{i,\text{acc}}+\beta_f\,\mathcal{S}_{i,\text{format}}\right)$\;
\Return $\mathcal{R}_{i,\text{final}}$\;
\end{algorithm}
\clearpage
\begin{algorithm}[h]
\caption{ResT: Reshaping Token-level Policy Gradients}
\label{alg:rest}
\DontPrintSemicolon
\KwIn{Prompt $x$; policy $\pi_\theta$ with frozen reference $\pi_{\theta_{\text{old}}}$; group size $G$; PPO clip $\epsilon$; stability constant $\delta>0$; curriculum proportion $\nu\in[0,1]$; clipping bounds $(w_{\min},w_{\max})$; step sizes $(\alpha_f,\alpha_p,\alpha_t)$; entropy-based initializer (Eq.~\ref{factor}).}
\KwOut{Updated parameters $\theta$.}
\BlankLine

\textbf{Rollout:}\;
\For{$i=1$ \KwTo $G$}{
  Sample response $y_{i}\sim \pi_{\theta}(\cdot\,|\,x)$ containing plain-text and tool-call segments\;
  Compute reward $\mathcal{R}_{i,\text{final}}$ via Alg.~\ref{alg:reward}\;
}

\textbf{Group-normalized advantage:}\;
$\mu \leftarrow \frac{1}{G}\sum_{i=1}^G \mathcal{R}_{i,\text{final}}$,\quad
$\sigma \leftarrow \sqrt{\frac{1}{G}\sum_{i=1}^G (\mathcal{R}_{i,\text{final}}-\mu)^2}+\delta$\;
\For{$i=1$ \KwTo $G$}{
  $\hat{A}_i \leftarrow \dfrac{\mathcal{R}_{i,\text{final}}-\mu}{\sigma}$\;
}

\textbf{Token partition and initialization:}\;
\tcp{Initialize global token weights (shared across samples)}
Let $T$ be a representative sequence length or maximum length\;
Initialize $\tilde w_t$ for all $t = 1,\ldots,T$ using entropy-based rule (Eq.~\ref{factor})\;

\textbf{Curriculum reweighting (global schedule)}:\;
\tcp{Apply curriculum updates once, shared by all sequences}
\For{$t=1$ \KwTo $T$}{
  $\tilde w_t \leftarrow \mathrm{Clip}(\tilde w_t, w_{\min}, w_{\max})$\;
}
\For{$t \in \mathcal{I}_{\mathrm{fmt}}$}{ $\tilde w_t \leftarrow \mathrm{Clip}(\tilde w_t - \alpha_f \nu,\; w_{\min}, w_{\max})$ }
\For{$t \in \mathcal{I}_{\mathrm{name}}$}{ $\tilde w_t \leftarrow w_{\max}$ }
\For{$t \in \mathcal{I}_{\mathrm{para}}$}{ $\tilde w_t \leftarrow \mathrm{Clip}(\tilde w_t + \alpha_p \nu,\; w_{\min}, w_{\max})$ }
\For{$t \in \mathcal{I}_{\mathrm{thk}}$}{ $\tilde w_t \leftarrow \mathrm{Clip}(\tilde w_t + \alpha_t \nu,\; w_{\min}, w_{\max})$ }

\tcp{Normalize globally}
$\bar w \leftarrow \frac{1}{T} \sum_{t=1}^{T} \tilde w_t$\;
\For{$t=1$ \KwTo $T$}{
  $\omega_t \leftarrow \dfrac{\tilde w_t}{\bar w + \delta}$\;
}

\textbf{PPO-style token-level objective and update:}\;
\For{$i=1$ \KwTo $G$}{
  \For{$t=1$ \KwTo $T^i$}{
    $r_{i,t} \leftarrow \dfrac{\pi_{\theta}(y_{i,t} \mid x, y_{i,<t})}{\pi_{\theta_{\text{old}}}(y_{i,t} \mid x, y_{i,<t})}$\;
    $g_{i,t} \leftarrow \min\!\Big( r_{i,t} \hat{A}_i,\; \mathrm{clip}(r_{i,t}, 1-\epsilon, 1+\epsilon) \hat{A}_i \Big)$\;
  }
}
$\mathcal{L}_{\text{ResT}}(\theta) \leftarrow -\frac{1}{G} \sum_{i=1}^{G} \sum_{t=1}^{T} -\frac{\omega_t}{T} \cdot g_{i,t}$\;
Update $\theta \leftarrow \theta - \eta\,\nabla_\theta \mathcal{L}_{\text{ResT}}(\theta)$\;
\end{algorithm}

\noindent\textbf{Helper:} $\mathrm{Clip}(u,a,b)=\min(\max(u,a),b)$.  Sets $\mathcal{I}_{fmt},\mathcal{I}_{name},\mathcal{I}_{para},\mathcal{I}_{thk}$ are obtained via deterministic token tagging of the generated sequence.  Eq.~\ref{factor} provides the entropy-based initializer for $\tilde w_t$.

\clearpage
\section{Complete Experiments Results}
\label{appendix-a}

\begin{figure}[h]
\centering
\setlength{\fboxsep}{0pt}
\includegraphics[width=0.95\linewidth]{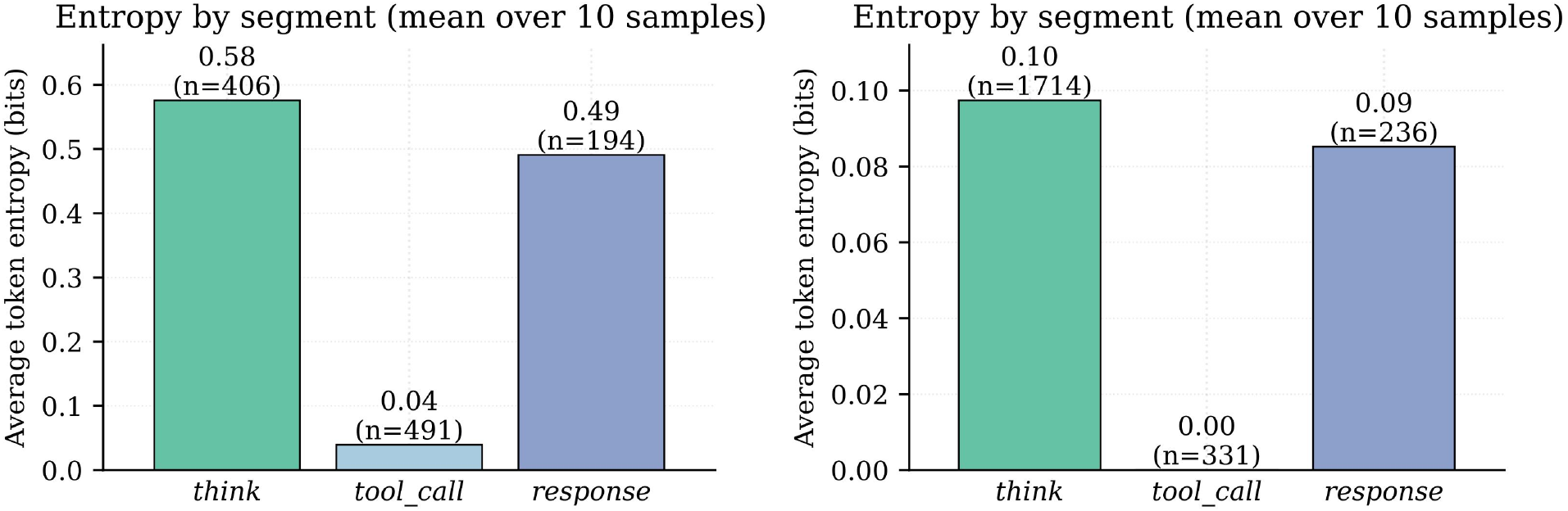}
\caption{Average region entropy on Llama-3.2-3B-Instruct (left) and Qwen3-1.7B (right). For each model, we randomly sample 10 instances and compute the entropy distribution over generated tokens. Tokens are partitioned into functional regions—reasoning, tool invocation, and final response.}
\label{fig:entropy}

\end{figure}

\begin{table}[h]
  \centering
  \caption{\textbf{API-Bank Test Results on Llama3.2-3B-Instruct model.} We report the highest accuracy achieved for each method after hyper-parameter tuning, with a breakdown by difficulty level.}
  \label{Llama}
  \resizebox{\linewidth}{!}{
  \begin{tabular}{lcccc}
\hline 
    \textbf{Model} & \textbf{Overall Acc} & \textbf{Level 1 Acc} & \textbf{Level 2 Acc} & \textbf{Level 3 Acc} \\
     \hline 
Llama-3.2-3B-Instruct(Base) & 37.35\% & 43.61\% & 32.84\% & 20.61\% \\
Llama-3.2-3B-Instruct(SFT) & 46.4\% & 52.88\% & 29.85\% & 35.11\% \\
Llama-3.2-3B-Instruct(TSFT) & 49.25\% & 57.14\% & 31.34\% & 34.35\% \\
Llama-3.2-3B-Instruct(RSFT) &51.26\% & 57.89\% & 29.85\% & 41.98\% \\
Llama-3.2-3B-Instruct(GRPO) & 59.13\% & 65.66\% & 52.24\% & 42.75\% \\
Llama-3.2-3B-Instruct(SFT+GRPO) & 52.76\% & 61.15\% & 43.28\% & 32.06\% \\
Llama-3.2-3B-Instruct(Ours, ResT)  & \textbf{62.81\%} & 69.42\% & 53.73\% & 47.33\% \\
 \hline
  \end{tabular}}

\end{table}
\clearpage
\begin{table}[h]
  \centering

  \caption{\textbf{BFCL Benchmark Results on Single Turn Scenarios (last updated June 14, 2025).} All metrics are calculated using the official scripts.}
  \label{bfcl-ablation}
 \resizebox{\linewidth}{!}{
  \begin{tabular}{lcccc}
\\ \hline \\
    \textbf{Model} & \textbf{Non-Live AST Acc} & \textbf{Live Acc} & \textbf{Relevance Detection} & \textbf{Irrelevance Detection} \\
\hline \\

    Qwen3-1.7B (Base) & 83.27\% & 72.95\% & 83.33\% & 85.49\%  \\
    Qwen3-1.7B (SFT) & 82.90\%	&74.06\% &72.22\% &78.02\%  \\
    Qwen3-1.7B (TSFT) & 83.67\% &74.59\%&77.78\%&78.04\%  \\
    Qwen3-1.7B (RSFT) & 82.46\%&	74.19\%&	66.67\%&	76.53\%  \\
    Qwen3-1.7B (GRPO) & 83.00\% & 72.95\% & 72.22\% & 73.42\%  \\
    Qwen3-1.7B (SFT+GRPO) & \textbf{85.31\%} & 72.10\% & 66.67\% & 76.94\% \\
    Qwen3-1.7B (Ours, ResT) & {84.65\%} & \textbf{73.43\%} & 66.67\% & 76.30\%  \\
 \hline\\
    Qwen3-4B-2507 (Base) & 87.98\% & 79.52\% & 77.78\% & 79.18\%  \\
    Qwen3-4B-2507 (SFT)& 89.04\%&	80.94\%&	83.33\%&	81.90\%  \\
    Qwen3-4B-2507 (TSFT)& 87.31\%	&80.63\%	&83.33\%	&81.39\% \\
    Qwen3-4B-2507 (RSFT)& 87.69\%&	80.59\%&	88.89\%&	82.02\%  \\
    Qwen3-4B-2507 (GRPO) & {89.40}\% & 79.08\% & 83.33\% & 81.52\%  \\
    Qwen3-4B-2507 (SFT+GRPO) & \textbf{89.46}\% & 80.36\% & 77.78\% & 80.22\%  \\
    Qwen3-4B-2507 (Ours, ResT) & {88.65}\% & \textbf{81.12\%} & 77.78\% & 79.67\%  \\    
 \hline  \\
    Qwen3-8B (Base) & 88.81\% & 78.54\% & 77.78\% & 79.08\%  \\
    Qwen3-8B (SFT)& 82.58\%	&73.70\%	&72.22\%	&75.62\%  \\
    Qwen3-8B (TSFT)& 81.88\%&	75.26\%&	83.33\%	&76.72\%  \\
    Qwen3-8B (RSFT)& 81.08\%&	73.83\%&	77.78\%&	77.38\%  \\
    Qwen3-8B (GRPO) & 88.50\% & \textbf{80.85\%} & 83.33\% & 82.24\%  \\
    Qwen3-8B (SFT+GRPO) & 85.90\% & 75.70\% & 77.78\% & 75.96\%  \\
    Qwen3-8B (Ours, ResT) & \textbf{90.08\%} & 79.03\% & 94.44\% & 77.62\%  \\
\hline  \\
    Qwen3-14B (Base) & 88.17\% & 79.96\% & 72.22\% & 81.01\%  \\
    Qwen3-14B (SFT) & 77.31\%&	71.83\%	&77.78\%	&69.95\%  \\
    Qwen3-14B (TSFT) & 79.90\% & 71.97\% &	83.33\%&	72.48\%  \\
    Qwen3-14B (RSFT) & 71.81\% & 73.39\% & 55.56\% & 85.44\%  \\
    Qwen3-14B (GRPO) & 89.10\% & 79.74\% & 77.78\% & 79.31\%  \\
    Qwen3-14B (SFT+GRPO) & 86.56\% & 76.59\% & 77.78\% & 75.41\%  \\
    Qwen3-14B (Ours, ResT) & \textbf{89.10\%} & \textbf{80.81\%} & 77.78\% & 80.56\%  \\
 \hline 
  \end{tabular}}
\end{table}
\clearpage
\section{Group Relative Policy Optimization (GRPO)}

Group Relative Policy Optimization (GRPO) serves as a critic-free, on-policy optimization algorithm designed for Large Language Models (LLMs). It presents an efficient alternative to Proximal Policy Optimization (PPO) by constructing a relative baseline for advantage estimation. This approach circumvents the need to learn a separate value network, thereby avoiding dependencies on methods like Generalized Advantage Estimation (GAE) which require a value function.

The core idea of GRPO is to dynamically construct a baseline for each sampled output using the reward distribution of a set of candidate outputs generated from the same input prompt. Specifically, for a given prompt $q$, we first sample a set of $S$ complete output sequences $\mathcal{O} = \{o_1, o_2, \dots, o_S\}$ from the current policy $\pi_{\theta_{old}}$. A scalar reward $r_i = R(q, o_i)$ is then computed for each output $o_i$.

The empirical mean of these rewards serves as the baseline for evaluating each output, i.e., $b(q) = \text{mean}(\{r_1, \dots, r_S\})$. The advantage for each rollout $o_i$ is then normalized using the standard deviation of the rewards to obtain the normalized advantage estimate $\hat{A}_i$:
\begin{equation}
\label{eq:grpo_advantage}
\hat{A}_i = \frac{r_i - \text{mean}(\{r_1, \dots, r_S\})}{\text{std}(\{r_1, \dots, r_S\}) + \delta}
\end{equation}
where $\delta$ is a small constant added for numerical stability to prevent division by zero. This "intra-group relative" advantage estimation effectively reduces reward variance.

During the policy update phase, this sequence-level advantage value $\hat{A}_i$ is applied to every timestep (token) within that sequence. GRPO performs policy updates by maximizing a clipped surrogate objective, analogous to PPO:
\begin{align}
\mathcal{L}^{\text{GRPO}}(\theta)
&= \mathbb{E}_{q \sim \mathcal{D},\, o \sim \pi_{\theta_{\mathrm{old}}}}
   \left[
   \sum_{t=1}^{|o|}
   \min\!\Bigl(
       \rho_t(\theta)\hat{A},\;
       \text{clip}\!\bigl(\rho_t(\theta),\, 1\!-\!\epsilon,\, 1\!+\!\epsilon\bigr)\hat{A}
   \Bigr)
   \right. \notag\\
&\qquad\qquad\quad
   \left.
   -\; \beta\,\text{KL}\!\bigl(\pi_{\theta}(\cdot|q)\big\| \pi_{\text{ref}}(\cdot|q)\bigr)
   \right].
\label{eq:grpo_objective}
\end{align}
where $\rho_t(\theta) = \frac{\pi_\theta(o_t | q, o_{<t})}{\pi_{\theta_{old}}(o_t | q, o_{<t})}$ is the likelihood ratio of the new and old policies at timestep $t$. $\hat{A}$ corresponds to the normalized advantage of the current sequence $o$. The hyperparameters $\epsilon$ and $\beta$ control the clipping range and the strength of the KL-divergence penalty against a reference policy $\pi_{\text{ref}}$, respectively.

By adopting this formulation, GRPO retains the stability of PPO's trust-region updates while eliminating the complexities of training a value network, offering a simple yet effective framework for fine-tuning LLMs with reinforcement learning.
\end{document}

%% file: math_commands.tex
\usepackage{amsmath,amsfonts,bm}

\def\eqref#1{equation~\ref{#1}}

\def\1{\bm{1}}

\DeclareMathAlphabet{\mathsfit}{\encodingdefault}{\sfdefault}{m}{sl}
\SetMathAlphabet{\mathsfit}{bold}{\encodingdefault}{\sfdefault}{bx}{n}

\newcommand{\E}{\mathbb{E}}

\newcommand{\softmax}{\mathrm{softmax}}

\newcommand{\Var}{\mathrm{Var}}

\newcommand{\Cov}{\mathrm{Cov}}
